\documentclass{bmvc2k}

\usepackage{needspace}
\usepackage{kotex}
\usepackage{amsmath}
\usepackage{amssymb}
\usepackage{booktabs}
\usepackage{multirow}
\usepackage{makecell}
\usepackage{graphicx}
\usepackage{pifont}
\usepackage{comment}

\usepackage{subcaption}
\definecolor{bmvcaptioncolor}{rgb}{0,0,0.4}
\DeclareCaptionFont{color}{\color{bmvcaptioncolor}}

\title{SMART: MLLM-guided Temporal Alignment for Unifying Sign Language Recognition and Spotting}

\addauthor{Eunjee Choi$^{*}$}{eun09ji@dankook.ac.kr}{1}
\addauthor{JungHoon Sung$^{*}$}{jason_dku77@dankook.ac.kr}{1}
\addauthor{Seongwhan Cho}{chosw0812@dankook.ac.kr}{1}
\addauthor{Chu Xin}{chuxin@dankook.ac.kr}{1}
\addauthor{Younggeun Choi}{younggch@dankook.ac.kr}{1}

\addinstitution{
 Dankook University\\
 Yongin, Korea
}

\runninghead{E. CHOI, J. SUNG, S. CHO, X. CHU, Y. CHOI}{SMART}

\begin{document}

\maketitle

\begingroup
\renewcommand{\thefootnote}{\fnsymbol{footnote}}
\footnotetext[1]{Equal contribution.}
\endgroup

\begin{abstract} 
Continuous sign language recognition (CSLR) aims to recognize gloss sequences from unsegmented sign videos under weak sequence-level supervision. However, existing methods rely on sentence-level gloss annotations, providing limited temporal and semantic guidance for fine-grained representation learning. Conventional video-text alignment also requires large batch sizes, making it inefficient for memory-intensive sign language video training. In this work, we propose \textbf{SMART}, an MLLM-guided temporal alignment framework for joint sign recognition and spotting. \textbf{SMART} uses MLLM-generated motion descriptions as auxiliary semantic cues and performs stable video-text alignment under small-batch training. To improve temporal representation learning, we introduce a Multi-Scale Temporal Adapter that models temporal interactions during transformer encoding. For dense temporal localization, \textbf{SMART} incorporates CSFormer, a CSLR-guided spotting module that injects recognition-derived gloss evidence into a boundary-aware spotting network. This unified framework enables CSLR features to benefit spotting, while spotting supervision complements weak CTC-based recognition. Experiments on four sign language benchmarks, including PHOENIX14-T, CSL-Daily, Large-scale KSL, and Disaster and Safety KSL datasets, demonstrate the effectiveness of \textbf{SMART} across both recognition and spotting tasks.


\end{abstract}     

\section{Introduction}
\label{sec:intro}

\begin{figure*}[t]
    \centering
    \includegraphics[width=0.9\textwidth]{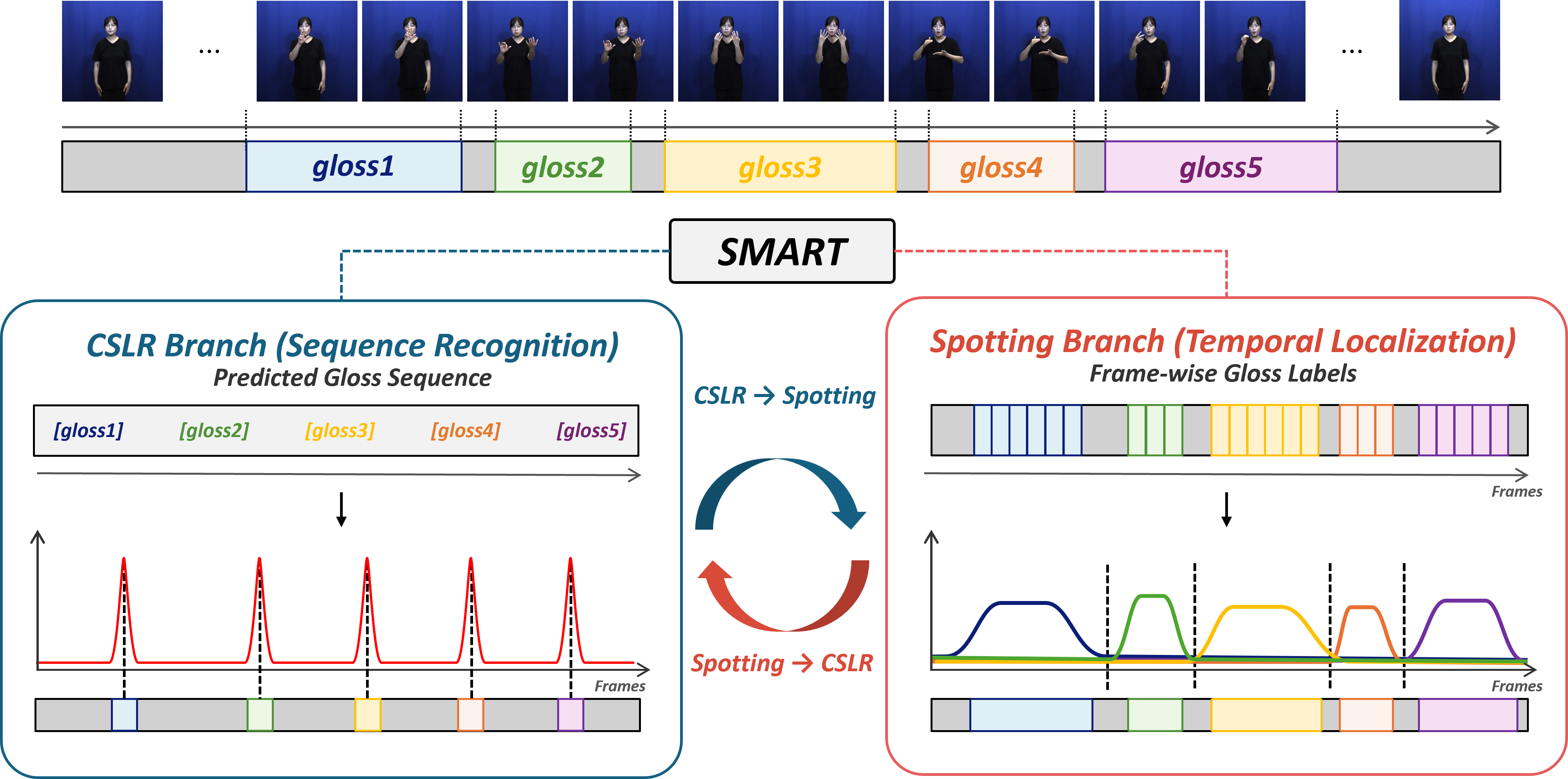}
    \vspace{3mm}
    \caption{\textbf{SMART} unifies sign language recognition and spotting through complementary recognition and localization supervision. The spotting branch refines weak CTC-based temporal alignments with dense temporal supervision, while the CSLR branch provides recognition-derived gloss evidence for spotting.
}
    \label{fig:teaser}
\end{figure*}

Sign language is a main communication medium for deaf and hard-of-hearing people. Developing effective sign language understanding systems is important for reducing communication barriers between signers and non-signers. Sign language understanding includes several tasks, such as sign language recognition, sign language translation, and sign spotting. In this work, we focus on Continuous Sign Language Recognition (CSLR) and sign spotting, which aim to recognize gloss sequences and localize signs in continuous videos, respectively.

CSLR recognizes gloss sequences from unsegmented sign videos using only sentence-level gloss annotations. Since frame-level gloss annotations are not available in CSLR, most CSLR methods rely on Connectionist Temporal Classification (CTC) loss ~\cite{graves2006connectionist, cui2019deep, cheng2020fully, cui2017recurrent}. Although CTC enables training without frame-level labels, it often produces peaky alignments~\cite{sjh_peaky_liu_nips2018, cui2019deep, pu2019iterative}, where each gloss is predicted by only a few frames while most frames are assigned to blank class. This weak temporal supervision limits dense frame-level representation learning and makes it difficult to capture fine-grained sign dynamics. 
Sign spotting can complement this limitation because it requires frame-level localization of target signs. As shown in Fig.~\ref{fig:teaser}, we integrate CSLR and spotting into a unified framework. We introduce CSFormer, a boundary-aware temporal segmentation backbone adapted for sign spotting, which refines temporal boundaries and provides dense frame-level predictions. Prior sign spotting methods have mainly been formulated as retrieval or localization tasks using short query templates or external sign dictionaries~\cite{sjh-read-spottin21, sjh-spotting_scalingup2022, sjh-spotting_watch2020, sjh-spotting_camgoz2021, eccv2022_challenge, eccv2022_hierarchical, sjh_2021_seg_&_spotting}, or have relied on sparsely annotated large-scale resources such as BSL-1K~\cite{sjh-bsl1k-albanie20}. Thus, dense frame-level sign spotting with large vocabularies and accurate boundary localization remains underexplored. To address this, we introduce a CSLR-aware injection mechanism that explicitly feeds gloss probabilities learned by the recognition module into the spotting network to guide temporal boundary estimation. Consequently, we demonstrate that CSLR features are highly effective for spotting segmentation, and conversely, the dense temporal supervision from the spotting module complements weak CTC-based recognition.

Recent CSLR methods have improved visual representation learning by adapting large-scale vision-language models. In particular, Adaptsign~\cite{hu2024improving} adopts a frozen CLIP~\cite{radford2021learning} backbone with lightweight adaptation module, showing that pretrained CLIP features can be efficiently transferred to CSLR. However, existing CLIP-based CSLR methods use CLIP as a visual feature extractor and still rely on CTC-based gloss supervision, without using language-based semantic guidance. 
To address these limitations, we use MLLM-generated motion descriptions as auxiliary semantic cues. Since sign language is expressed through hand movements, facial expressions, and body motions, these descriptions provide motion-aware language guidance for representation learning~\cite{kim2025leveraging}. We align video and text representations using a SigLIP~\cite{zhai2023sigmoid}, which enables stable video-text alignment under small-batch video training. To further improve temporal representation learning, we introduce a Multi-Scale Temporal Adapter (MSTA) inside the CLIP backbone. Although the CLIP image encoder provides image representations, it does not model temporal dependencies across frames. MSTA addresses this limitation by capturing multi-scale temporal interactions during transformer encoding while preserving the pretrained CLIP representations. In addition, we incorporate sign spotting as dense temporal supervision to complement weak CTC-based alignments. Therefore, we introduce CSFormer, a CSLR-aware spotting module that injects gloss evidence from the recognition branch into a boundary-aware temporal segmentation backbone, enabling fine-grained boundary estimation for sign spotting. Based on these reasons, we propose \textbf{SMART}, an MLLM-guided Temporal Alignment framework for CSLR and spotting. Our contributions are summarized as follows:

\begin{itemize} 
    \item We propose \textbf{SMART}, an MLLM-guided temporal alignment framework for CSLR and spotting. \textbf{SMART} uses  MLLM-generated motion descriptions as auxiliary semantic cues and enables stable video-text alignment under small-batch training.

    \item A Multi-Scale Temporal Adapter is introduced to capture multi-scale temporal dependencies within the CLIP backbone for enhanced temporal representation learning.

    \item We jointly integrate sign spotting and CSLR via \textbf{CSFormer}, using recognition-derived gloss guidance for dense frame-level localization and late fusion to complement weak CTC-based recognition.

    \item We conduct experiments on four sign language benchmarks, including PHOENIX14-T, CSL-Daily, Large-scale KSL, and Disaster and Safety KSL. The two Korean sign language datasets are used for CSLR and sign spotting for the first time, and the results demonstrate the complementarity between gloss-level recognition guidance and dense temporal supervision across different sign languages.

\end{itemize}

\section{Related Works}
\label{sec:related_works}

\textbf{Continuous Sign Language Recognition.}
Continuous Sign Language Recognition (CSLR) aims to recognize gloss sequences from unsegmented sign videos, where only sentence-level gloss annotations are available during training~\cite{cui2019deep, koller2019weakly}. Early CSLR approaches mainly relied on hand-crafted visual features~\cite{freeman1995orientation, koller2015continuous, sun2013discriminative} or HMM-based recognition systems~\cite{gao2004chinese, han2009modelling, ong2005automatic} for temporal modeling. With the development of deep learning, CTC loss~\cite{graves2006connectionist} has been widely adopted for end-to-end CSLR~\cite{cihan2017subunets, cui2017recurrent}. Most CTC-based methods consist of a spatial backbone for frame-wise feature extraction and a temporal module, such as a 1D CNN, LSTM, or transformer, for sequence modeling~\cite{cihan2017subunets, cui2017recurrent, cheng2020fully, niu2020stochastic}. Since CSLR datasets provide only sentence-level gloss annotations without frame-level gloss boundaries or temporal alignments, CTC-based methods receive only weak temporal supervision. As a result, CTC often produces peaky alignments~\cite{sjh_peaky_liu_nips2018}, where each gloss is predicted by only a few frames while most frames are assigned to the blank class. Although effective for sentence-level recognition, such alignments provide limited supervision for modeling fine-grained temporal dynamics. Recent methods have advanced CSLR by improving visual representations~\cite{cheng2020fully}, refining temporal modeling~\cite{niu2020stochastic}, or adapting pretrained vision-language models. In particular, AdaptSign~\cite{hu2024improving} adopts a frozen CLIP~\cite{radford2021learning} backbone and introduces lightweight adaptation modules to transfer pretrained visual representations to CSLR. This shows the potential of adapting large-scale pretrained vision-language models for CSLR. However, existing CSLR methods still rely on sequence-level gloss supervision, while the use of MLLM-generated language descriptions as auxiliary semantic cues for CSLR remains underexplored.

\noindent\textbf{Peaky Alignment in CSLR.} 
Due to the absence of ground-truth frame-level alignment, the Connectionist Temporal Classification (CTC) loss is widely adopted as the standard objective~\cite{graves2006connectionist, camgoz2018neural}. As a side effect of this formulation, models typically suffer from the \textit{peaky alignment} problem; they predict each gloss as a single time-step spike within the downsampled feature space while filling the remaining sequence with blank (background) tokens~\cite{sjh_peaky_liu_nips2018, sjh_peaky_zeyer_does2021}. Although this peakiness is not reflected in sentence-level Word Error Rate (WER) scores alone, it poses an inherent limitation for CTC-trained CSLR models~\cite{min2021visual, zuo2022c2slr, hao2021self}. Because downsampled feature predictions are restricted to a combination of blanks and isolated spikes, the model has difficulty capturing the full temporal structure of each sign. Furthermore, the lack of meaningful gradient flow to adjacent frames significantly prevents the model's ability to capture the intrinsic temporal consistency of each sign. To overcome this limitation, we propose a framework that integrates CSFormer to provide dense, frame-level predictions. Consequently, we demonstrate that our approach alleviates the frame-level peaky limitation and further complements sentence-level recognition.

\noindent\textbf{Action segmentation and Sign Spotting.}
Frame-level action segmentation is a representative task requiring dense temporal modeling, making it the field most closely related to sign spotting. MS-TCN~\cite{sjh-ms_tcn-farha2019ms} proposes a multi-stage temporal convolutional network architecture that progressively refines frame-wise class probabilities at each stage, which is subsequently advanced by MS-TCN++~\cite{sjh-ms_tcn-li2023ms} to enhance dilated information exchange across stages. Among transformer-based approaches, ASFormer~\cite{sjh-asformer21} substantially boosts segmentation accuracy by combining sliding-window self-attention with a multi-stage decoder. Additionally, ASRF~\cite{sjh-asrf21} introduces a refinement module equipped with a dedicated boundary regression head to post-process and refine the temporal boundaries of predicted segments. More recently, LTContext~\cite{sjh-ltcontext23} effectively captures long-range dependencies in long-duration videos by disentangling windowed attention from long-term attention. However, these methods rely on visual features from pre-trained models, or optical flow as inputs, and the effectiveness of backbones designed for the sign language domain has yet to be investigated.
Sign spotting is a recently introduced task that aims to predict the temporal intervals in which signs from a predefined vocabulary occur in a video. However, prior sign spotting research has primarily focused on retrieval-based formulations that rely on short query templates~\cite{sjh-read-spottin21, sjh-spotting_scalingup2022, sjh-spotting_watch2020, sjh-spotting_camgoz2021, eccv2022_challenge, eccv2022_hierarchical, sjh_2021_seg_&_spotting} or on sparsely annotated datasets such as BSL-1K~\cite{sjh-bsl1k-albanie20}. Consequently, sign spotting formulated as a dense, frame-level recognition task under large vocabularies and abundant training data remains largely unexplored.

\section{Methods}
\label{sec:methods}

\begin{figure*}[t]
    \centering
    \includegraphics[width=\textwidth]{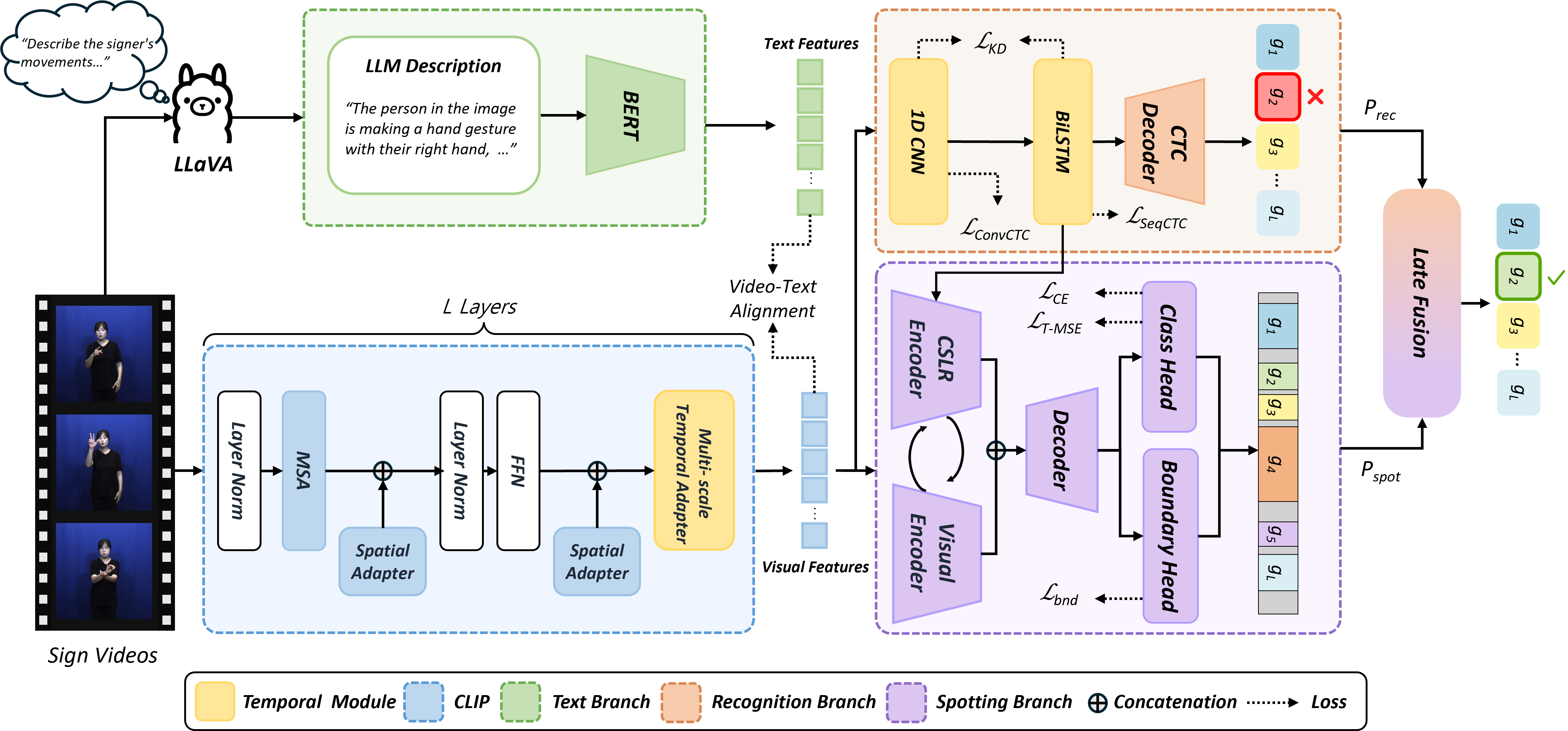}
    \caption{ 
    Overview of the proposed \textbf{SMART} framework. Given an input sign video, the visual encoder extracts spatio-temporal representations enhanced by the proposed Multi-Scale Temporal Adapter. MLLM-generated motion descriptions provide auxiliary semantic cues through video-text alignment. In addition, the CSLR-guided sign spotting branch provides temporal supervision for boundary refinement, enabling joint optimization of CSLR and sign spotting. 
    } 
    \label{fig:overview}
\end{figure*}

\begin{figure*}[t]
    \centering
    \newlength{\figboxheight}
    \setlength{\figboxheight}{0.25\textheight}

    \makebox[\textwidth][c]{%
    \begin{minipage}[t]{0.45\textwidth}
        \centering
        \includegraphics[
            width=\linewidth,
            height=\figboxheight,
            keepaspectratio
        ]{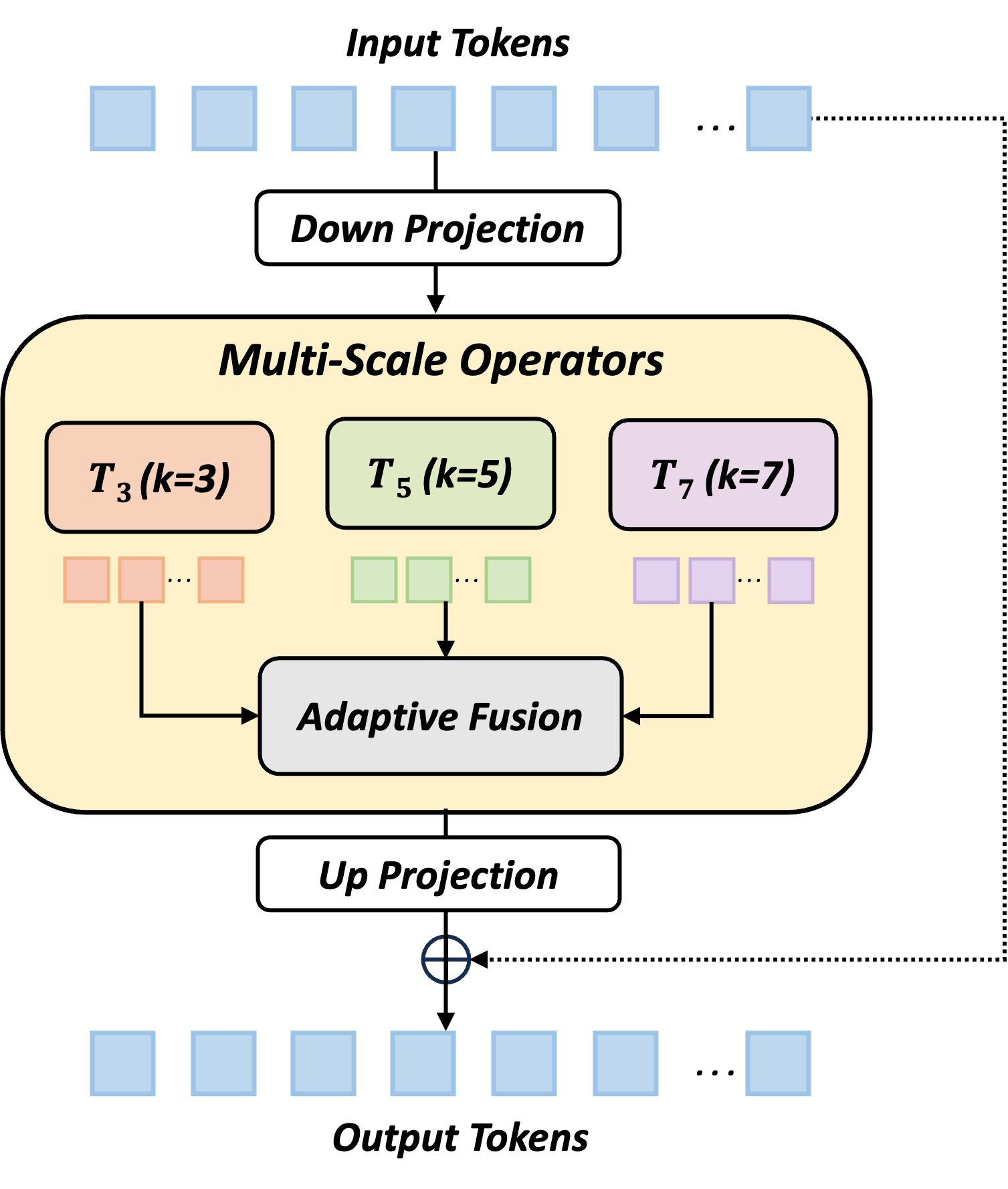}
        \par\vspace{1mm}
        {\small (a) Multi-Scale Temporal Adapter module.}
    \end{minipage}
    \hspace{0.04\textwidth}
    \begin{minipage}[t]{0.45\textwidth}
        \centering
        \includegraphics[
            width=\linewidth,
            height=\figboxheight,
            keepaspectratio
        ]{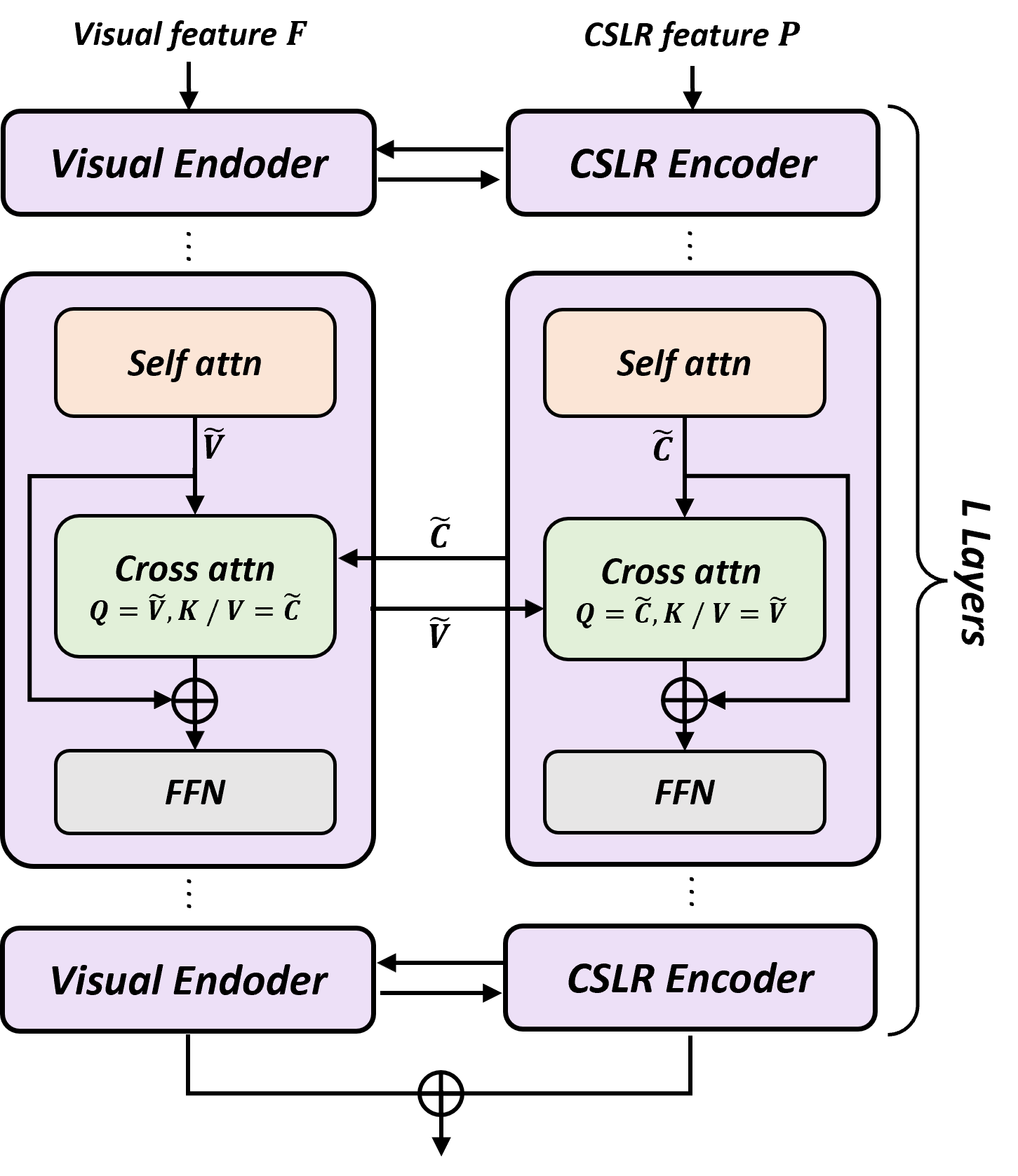}
        \par\vspace{1mm}
        {\small (b) CSFormer sign spotting module.}
    \end{minipage}
    }
    \caption{Illustration of the proposed Multi-Scale Temporal Adapter and CSFormer sign spotting module.}
    \label{fig:msta_spotting}
\end{figure*}

We propose \textbf{SMART}: MLLM-guided Temporal Alignment for Sign Recognition and Spotting. \textbf{SMART} consists of three key components: (1) an MLLM-guided video-text alignment module that aggregates frame-level motion descriptions into video-level semantic cues, (2) a video encoder enhanced by the proposed Multi-Scale Temporal Adapter (MSTA) for spatio-temporal feature extraction, and (3) a CSLR-aware sign spotting module, CSFormer, that injects recognition probabilities into a boundary-aware backbone for fine-grained temporal refinement. As illustrated in Fig.~\ref{fig:overview}, these components are jointly modeled for Continuous Sign Language Recognition (CSLR) and sign spotting. Each component is described in Sec.~\ref{sec:mllm}--\ref{sec:spotting}. 

\subsection{MLLM Frame-Level Text Generation}
\label{sec:mllm}
Prior work~\cite{kim2025leveraging} uses MLLM-generated sign descriptions for sign language translation. We extend this idea to CSLR by generating frame-level motion descriptions and using them as auxiliary semantic cues for video-text alignment. Before training, each sign video is processed with LLaVA-OneVision-7B~\cite{li2024llava} to generate descriptions focusing on the signer's hand movements and facial expressions.

\subsection{Sign Language Recognition}
\label{sec:recognition} 
The \textbf{SMART} architecture consists of a CLIP ViT-B/16 visual encoder~\cite{radford2021learning, dosovitskiy2020image}, which is augmented with Spatial Adapters and Prefix embeddings~\cite{hu2024improving}, and the proposed MSTA. The extracted visual features are then fed into a sequence model composed of a 1D CNN and a two-layer bidirectional LSTM. The final features are passed to a NormLinear classifier and CTC decoder for gloss sequence prediction. For semantic supervision, video representations are aligned with pre-extracted BERT text embeddings via video-text alignment during training.

\noindent\textbf{Visual Encoder.}
Given a sign language video with $T$ frames $\mathbf{x} = \{x_t\}_{t=1}^{T} \in \mathbb{R}^{T \times 3 \times 256 \times 256}$, each frame is processed by a frozen CLIP ViT-B/16 image encoder. Spatial Adapters and Prefix embeddings~\cite{hu2024improving} are incorporated into each transformer block for spatial adaptation to the sign language domain. However, since CLIP processes each frame independently, temporal dynamics across frames are not captured within the encoder. To address this, we propose the MSTA, inserted into the last 4 transformer blocks to model inter-frame temporal dependencies at the feature level. The features are then passed through a 1D CNN and a two-layer bidirectional LSTM, followed by a NormLinear Classifier and CTC decoder for gloss sequence prediction. 

\noindent\textit{Multi-Scale Temporal Adapter.}
The architecture of MSTA is illustrated in Fig.~\ref{fig:msta_spotting}(a). Formally, given visual token features $\mathbf{Z} \in \mathbb{R}^{L \times BT \times d}$, where $L$, $B$, $T$, and $d$ denote the number of visual tokens, batch size, number of frames, and feature dimension, respectively, MSTA first projects $\mathbf{Z}$ into a low-dimensional temporal adaptation space:
\begin{equation}
\mathbf{h} = \phi(\mathrm{LN}(\mathbf{Z}\mathbf{W}_{down})),
\end{equation}
where $\mathbf{h} \in \mathbb{R}^{L \times BT \times d_m}$ and $d_m=d/r$. The feature $\mathbf{h}$ is then reshaped along the temporal dimension, and multi-scale temporal operators are applied:
\begin{equation}
\hat{\mathbf{h}} =
\sum_{k \in \{3,5,7\}} \alpha_k \mathcal{T}_k(\mathbf{h}),
\qquad
\boldsymbol{\alpha}=\mathrm{softmax}(\boldsymbol{\theta}),
\end{equation}
where $\mathcal{T}_k(\cdot)$ denotes a temporal operator with kernel size $k$, and the fusion weights $\boldsymbol{\alpha}$ are learned through a softmax-normalized learnable parameter vector $\boldsymbol{\theta}\in\mathbb{R}^{3}$. The multi-scale temporal operators capture motion patterns with different temporal receptive fields, enabling the model to encode both short- and long-range temporal dependencies.

The adapted temporal feature is projected back to the original feature dimension and added to the input through a residual connection:
\begin{equation}
\mathbf{Z}' = \mathbf{Z} + \hat{\mathbf{h}}\mathbf{W}_{up}.
\end{equation}
The up-projection $\mathbf{W}_{up}$ is zero-initialized to preserve the pretrained CLIP representations at the beginning of training.

\noindent\textbf{Text Encoder.}
The generated frame-level motion descriptions are encoded using BERT~\cite{devlin2019bert}. For each frame description, the final-layer representation of the \texttt{[CLS]} token is extracted as a $768$-dimensional text embedding, resulting in per-frame text embeddings $\mathbf{T} \in \mathbb{R}^{F \times 768}$, where $F$ denotes the number of sampled frames. These embeddings are averaged across frames to produce a single global text embedding $\mathbf{t} \in \mathbb{R}^{768}$ for each video, which is used for video-text alignment during training.

\noindent\textbf{Video-Text Alignment.}
Existing CSLR models rely solely on CTC loss, which provides only sentence-level supervision. MLLM-generated motion descriptions provide semantic information about hand movements and facial expressions, but directly fusing text features into the visual encoder risks degrading the pretrained CLIP representations and increases inference cost. Instead, we use text embeddings as auxiliary supervision through video-text alignment, which bridges the semantic gap between visual and textual modality during training. InfoNCE losses~\cite{radford2021learning} rely on in-batch negative pairs, requiring large batch sizes to provide sufficient negative samples. This setting is inefficient for memory-intensive CSLR training, where batch sizes are small. We instead adopt SigLIP~\cite{zhai2023sigmoid}, which applies a sigmoid loss independently to each pair, enabling stable optimization in small batch settings.

Specifically, after the CLIP visual encoder extracts frame-wise features, we apply temporal pooling over frames to obtain a video-level representation, which is then projected to a $768$-dimensional video embedding $\mathbf{v} \in \mathbb{R}^{768}$ via a video projector. The global text embedding $\mathbf{t} \in \mathbb{R}^{768}$ from Sec.~\ref{sec:mllm} is projected via a text projector. Both embeddings are $\ell_2$-normalized before alignment. The alignment loss is defined as:
\begin{equation} 
\mathcal{L}_{\text{Align}} = -\frac{1}{B^2}
\sum_{i=1}^{B} \sum_{j=1}^{B}
\log \sigma \bigl( y_{ij} \cdot s \cdot 
\mathbf{v}_i^\top \mathbf{t}_j \bigr),
\end{equation}
where $B$ denotes the batch size, $\mathbf{v}_i$ and $\mathbf{t}_j$ represent the normalized video and text embeddings of the $i$-th and $j$-th samples, respectively. The pair label $y_{ij}$ is set to $+1$ for matched video-text pairs and $-1$ otherwise, $\sigma(\cdot)$ is the sigmoid function, and $s$ is a learnable logit scale.

\noindent\textbf{Training Details.}
The total training loss combines CTC-based recognition losses, a knowledge distillation loss, and the proposed video-text alignment loss:
\begin{equation}
\mathcal{L}_{\text{total}} = 
\lambda_{\text{seq}}\mathcal{L}_{\text{SeqCTC}} + 
\lambda_{\text{conv}}\mathcal{L}_{\text{ConvCTC}} + 
\lambda_{\text{kd}}\mathcal{L}_{\text{KD}} + 
\lambda_{\text{align}}\mathcal{L}_{\text{Align}},
\end{equation}
where $\mathcal{L}_{\text{ConvCTC}}$ and $\mathcal{L}_{\text{SeqCTC}}$ are CTC losses applied to the 1D CNN and BiLSTM outputs respectively, and $\mathcal{L}_{\text{KD}}$ is a knowledge distillation loss between the two predictions~\cite{min2021visual,hu2024improving}. The weights are set to $\lambda_{\text{seq}}=1.0$, $\lambda_{\text{conv}}=1.0$, $\lambda_{\text{kd}}=25.0$, and $\lambda_{\text{align}}=0.05$. During inference, the alignment module is discarded, and recognition is performed using CTC beam search on the final BiLSTM logits to obtain the gloss sequence $\hat{\mathbf{y}} = \{\hat{y}_n\}_{n=1}^{N}$, where $N$ denotes the number of predicted glosses.

\subsection{Sign Language Spotting}
\label{sec:spotting}
We inject the gloss-sequence probabilities produced by the recognition module in Sec.~\ref{sec:recognition} into our spotting model, so that recognition-derived gloss evidence can guide the temporal boundary estimation in spotting. Existing sign spotting models predict gloss boundaries from visual features, which limits precision on short or visually ambiguous signs. However, CSLR predictions provide sparse but discriminative cues as peaky CTC spikes around gloss positions, whereas the visual encoder learns temporally continuous cues. Therefore, a naive channel-wise concatenation followed by a shared encoder can cause the sparse CSLR signal to be averaged out at the input stage. Built on the backbone~\cite{sjh-asformer21}, our model introduces three key modules: (1) a CSLR-aware injection that projects the visual features and the CSLR sequence logits into two independent streams, (2) a layer-wise bidirectional cross-attention that exchanges information between the two streams at every encoder layer, and (3) a boundary head on every stage that predicts gloss transitions jointly with frame-level classification. The overall structure is illustrated in Fig.~\ref{fig:msta_spotting}(b).

\noindent\textbf{CSLR-aware injection.}
Given frame-wise visual features $\mathbf{F} \in \mathbb{R}^{T \times 512}$ and the sequence logits $\mathbf{P} \in \mathbb{R}^{T_c \times C}$ produced by the BiLSTM classifier of Sec.~\ref{sec:recognition} ($C$: gloss vocabulary size), we first align $\mathbf{P}$ to length $T$ via 1D linear interpolation to obtain $\bar{\mathbf{P}} \in \mathbb{R}^{T \times C}$. Each modality is projected to the encoder dimension $D$ through a 1D convolution and kept as an independent stream throughout the encoder: 
\begin{equation}
\mathbf{V}^{(0)} = \text{Conv}_{1\times1}(\mathbf{F}), \quad
\mathbf{C}^{(0)} = \text{Conv}_{1\times1}(\bar{\mathbf{P}}).
\end{equation}

The recognition module is frozen during spotting training, so the spotting model receives recognition-derived gloss evidence only through $\bar{\mathbf{P}}$. At each encoder layer $l$, the two streams are first updated by sliding-window self-attention, then exchange information through a bidirectional cross-attention block. The visual stream attends to the peak locations of the CSLR predictions through the attention weights, while the CSLR stream refines its sparse spikes using the temporally continuous cues from the visual features. After $L$ layers, the final encoder output is defined as the concatenation of the two streams, which is forwarded to the decoder stages.

\noindent\textbf{Boundary Head and Multi-Task Objective.}
To explicitly model gloss transitions, we attach a boundary head implemented as a 1D convolution on top of the last feature of every stage. The boundary targets $b_t \in \{0,1\}$ are derived from frame-level GT labels by marking frames where the class index differs from the previous valid frame. To address the heavy positive/negative imbalance, $\mathcal{L}_{\text{bnd}}$ is defined as a weighted binary cross-entropy:
\begin{equation}
\mathcal{L}_{\text{spot}} = \mathcal{L}_{\text{ce}} + \lambda_{\text{mse}}\mathcal{L}_{\text{T-MSE}} + \lambda_{\text{bnd}}\mathcal{L}_{\text{bnd}}.
\end{equation}
The weights are set to $\lambda_{\text{mse}}=0.15$ and $\lambda_{\text{bnd}}=0.5$ in all experiments.

\noindent\textbf{Inference and Late Fusion.}
At inference, the classification logits from the last decoder stage are used for frame-level gloss prediction. To exploit the complementarity between recognition and spotting, we fuse the gloss probabilities of the two models after temporal alignment via a simple linear combination, followed by CTC beam-search decoding: 
\begin{equation}
\mathbf{P}_{\text{fuse}} = \alpha\,\mathbf{P}_{\text{rec}} + (1-\alpha)\,\mathbf{P}_{\text{spot}}.
\end{equation}
We set $\alpha=0.7$ in our grid search experiments.

\section{Experiments}
\label{sec:experiments}

\subsection{Experimental Settings}
\begin{table*}[t]
\centering
\caption{Statistics of the datasets used in our experiments.}
\label{tab:dataset}

\small
\setlength{\tabcolsep}{3pt}

\begin{tabular}{ccccc}
\toprule

Dataset
& PHOENIX14-T
& CSL-Daily
& Large-scale KSL
& \makecell{DS KSL} \\ 

\midrule

Language
& German
& Chinese
& Korean
& Korean \\

Domain
& Weather
& Daily-life
& Daily-life
& Disaster(Weather) \\ 

\# Videos
& 8,257
& 20,654
& 35,987
& 28,250 \\

Split
& 7096 / 519 / 642
& 18401 / 1077 / 1176
& 25590 / 6397 / 4000
& 20089 / 5023 / 3138 \\

\# Glosses
& 1,066
& 2,000
& 440
& 2,496 \\

\# Signers
& 9
& 10
& 18
& 20 \\

\bottomrule
\end{tabular}
\end{table*}

\begin{figure*}[t]
    \centering
    \includegraphics[width=0.6\textwidth]{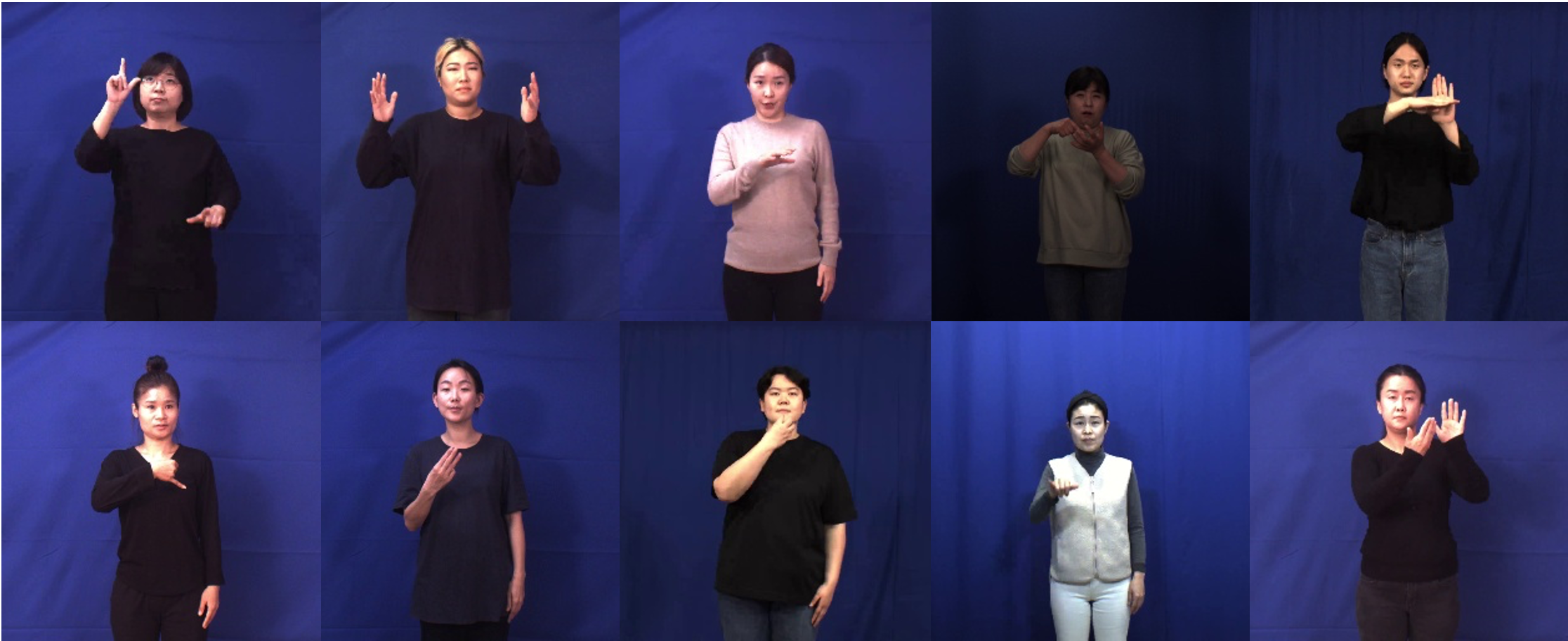}
    \caption{Examples of sign language videos from the DS KSL dataset.
    } 
    \label{fig:ksl}
\end{figure*}

\noindent\textbf{Datasets.}
Experiments are conducted on four continuous sign language recognition (CSLR) benchmarks: PHOENIX14-T~\cite{camgoz2018neural}, CSL-Daily~\cite{zhou2021improving}, the Large-scale KSL~\cite{aihub_ksl,ham2021ksl} and Disaster and Safety KSL (DS KSL)~\cite{aihub_disaster_ksl}. Dataset statistics are summarized in Tab.~\ref{tab:dataset}. PHOENIX14-T is a German sign language dataset collected from weather forecast broadcasts, while CSL-Daily consists of Chinese sign language videos covering daily-life scenarios. The two KSL datasets contain Korean sign language videos collected from daily-life and disaster domains. Unlike PHOENIX14-T and CSL-Daily, the KSL datasets provide temporal boundary annotations for sign spotting evaluation, enabling evaluation on both CSLR and sign spotting tasks. Representative frames from the DS KSL dataset are shown in Fig.~\ref{fig:ksl}. Since the KSL datasets only provide train and test splits, we further divide the training set into training and validation subsets using an 8:2 ratio.


\noindent\textbf{Implementation Details.} 
A CLIP-pretrained ViT-B/16 model~\cite{radford2021learning, dosovitskiy2020image} is used as the visual backbone. The model is trained for 40 epochs with a learning rate of $1\times10^{-4}$ and batch size 8. For video-text alignment, frame-level motion descriptions are generated using LLaVA-OneVision-7B~\cite{li2024llava} by processing videos, and encoded via BERT~\cite{devlin2019bert} to obtain text embeddings. The Multi-Scale Temporal Adapter (MSTA) is inserted into the last four CLIP transformer blocks with reduction factor $r=32$. For the sign spotting stage, CSFormer is built on an ASFormer~\cite{sjh-asformer21} encoder-decoder with $L{=}10$ encoder layers, a hidden dimension of $D{=}64$, and $3$ decoder stages. The recognition module is frozen during spotting training, and CSFormer is trained for 20 epochs with a batch size of 16 using Adam, a learning rate of $5\times10^{-4}$, and a 5-epoch warm-up. All experiments are conducted on a single NVIDIA RTX 6000 Ada GPU.

\noindent\textbf{Evaluation Metrics.} 
For CSLR, we use Word Error Rate (WER) as the evaluation metric. WER evaluates the difference between the predicted gloss sequence and the reference gloss sequence based on substitution, insertion, and deletion operations. Lower WER indicates better performance.
\begin{equation}
    \text{WER} = \frac{\#substitutions + \# insertions + \#deletions}{\#reference}
\end{equation}

For sign spotting, we report F1 score:
\begin{equation}
    \text{F1} = \frac{2 \cdot \text{Precision} \cdot \text{Recall}}{\text{Precision} + \text{Recall}}
\end{equation}
where a predicted segment is counted as a true positive if its temporal overlap with the ground truth exceeds a predefined IoU threshold. Higher F1 indicates better performance. We report F1 scores at varying IoU thresholds (0.10, 0.25, and 0.50) to simultaneously evaluate temporal localization and boundary alignment.

\begin{table*}[t]
\centering
\caption{Comparison with state-of-the-art methods on PHOENIX14-T, CSL-Daily, Large-scale KSL, and DS KSL datasets. Results are reported in WER, where lower is better. Here, \textbf{SMART (Ours)} denotes our recognition model without CSFormer. Best and second-best results are highlighted in bold and underline, respectively. $\dagger$ indicates that the baseline results on Large-scale KSL and DS KSL are reproduced using the official implementations under the original experimental settings.}
\label{tab:main_recognition}
\resizebox{\textwidth}{!}{
\begin{tabular}{lcccccccc}
\toprule
\multirow{2}{*}{Method} 
& \multicolumn{2}{c}{PHOENIX14-T} 
& \multicolumn{2}{c}{CSL-Daily} 
& \multicolumn{2}{c}{Large-scale KSL$^\dagger$} 
& \multicolumn{2}{c}{DS KSL$^\dagger$} \\
\cmidrule(lr){2-3}
\cmidrule(lr){4-5}
\cmidrule(lr){6-7}
\cmidrule(lr){8-9}
& Dev(\%) & Test(\%) & Dev(\%) & Test(\%) & Dev(\%) & Test(\%) & Dev(\%) & Test(\%) \\
\midrule

VAC~\cite{min2021visual} & - & - & - & - & 2.72 & 2.81 & 31.80 & 27.45 \\
SMKD~\cite{hao2021self} & 20.80 & 22.40 & - & - & - & - & - & - \\
TLP~\cite{hu2022temporal} & 19.40 & 21.20 & - & - & - & - & - & - \\
SEN~\cite{hu2023self} & 19.30 & 20.70 & - & - & 1.60 & 1.40 & 29.70 & 25.60 \\
AdaBrowse~\cite{hu2023adabrowse} & 19.50 & 20.60 & 31.20 & 30.70 & - & - & - & - \\
SSSLR~\cite{jang2023self} & 20.50 & 22.30 & - & - & - & - & - & - \\
C2SLR~\cite{zuo2022c2slr} & 20.20 & 20.40 & 31.90 & 31.00 & - & - & - & - \\
CoSign~\cite{jiao2023cosign} & 19.50 & 20.10 & 28.10 & 27.20 & - & - & - & - \\
SignBERTplus~\cite{hu2023signbert+} & 18.80 & 19.90 & - & - & - & - & - & - \\
CTCA~\cite{guo2023distilling} & 19.30 & 20.30 & 31.30 & 29.40 & - & - & - & - \\
CVT-SLR~\cite{zheng2023cvt} & 19.40 & 20.30 & - & - & - & - & - & - \\
CorrNet~\cite{hu2023continuous} & 18.90 & 20.05 & 30.60 & 30.10 & 2.62 & 2.64 & \underline{28.27} & \underline{25.79} \\
AdaptSign~\cite{hu2024improving} & \underline{18.60} & \underline{19.80} & \underline{26.70} & \underline{26.30} & \underline{0.89} & \underline{0.76} & 30.14 & 26.58 \\

\midrule
\textbf{SMART (Ours)} 
& \textbf{17.58} 
& \textbf{19.50} 
& \textbf{26.50} 
& \textbf{26.20} 
& \textbf{0.64} 
& \textbf{0.48} 
& \textbf{27.27} 
& \textbf{23.92} \\

\bottomrule
\end{tabular}
}
\end{table*}

~\begin{table*}[t]
\centering
\caption{Comparison with state-of-the-art methods on the Large-scale KSL and DS KSL benchmarks for sign spotting and spotting-enhanced recognition. For a fair comparison, all baseline methods are reproduced using the official implementations under the original experimental settings.}
\label{tab:spotting_main_both}
\scriptsize
\setlength{\tabcolsep}{2.7pt}
\resizebox{\textwidth}{!}{
\begin{tabular}{llccccc|ccccc}
\toprule
\multirow{2}{*}{Dataset} & \multirow{2}{*}{Method}
& \multicolumn{5}{c|}{Dev}
& \multicolumn{5}{c}{Test} \\
\cmidrule(lr){3-7} \cmidrule(lr){8-12}
& & F1@10 & F1@25 & F1@50 & Acc & WER(\%)& F1@10 & F1@25 & F1@50 & Acc & WER(\%)\\
\midrule

\multirow{8}{*}{LS KSL}
& SMART (CSLR-Only) & 35.23 & 18.88 & 7.47 & 59.05 & 0.64 & 36.58 & 19.59 & 7.37 & 59.64 & 0.48 \\
& HS-I3D~\cite{eccv2022_hierarchical}& 52.77& 33.01& 8.43& 57.79& -& 49.62& 30.84& 7.13& 57.37& -\\

& ASFormer~\cite{sjh-asformer21} & \underline{94.40}& \underline{94.31}& 89.39 & 88.93 & 6.27 & 94.30 & 94.25 & 89.32 & 88.58 & 6.69 \\
& ASRF~\cite{sjh-asrf21} & 91.23 & 90.04 & 75.05 & 81.11 & 12.89 & 88.09 & 86.78 & 72.18 & 80.38 & 16.24 \\
& LTContext~\cite{sjh-ltcontext23} & 93.32 & 93.18 & \underline{89.47}& \underline{89.47}& 6.79 & 92.55 & 92.40 & 86.58 & 87.54 & 9.27 \\
& VAC + CSFormer& 88.44 & 87.48 & 83.76 & 87.39 & 2.72 & 95.57 & 95.22 & 93.02 & 90.27 & 2.81 \\
& SEN + CSFormer& 90.68 & 90.14 & 87.64 & 89.35 & 1.60 & \underline{96.20}& \underline{95.99}& \underline{93.79}& 90.48 & 1.40 \\
& CorrNet + CSFormer& 91.45 & 90.95 & 88.33 & \underline{89.66}& 2.62 & 95.72 & 95.36 & 93.49 & 90.89 & 2.64 \\
& AdaptSign + CSFormer& 89.38 & 88.72 & 85.94 & 88.73 & \underline{0.89}& 95.64 & 95.23 & 93.19 & \underline{91.04}& \underline{0.76}\\
\cmidrule(lr){2-12}

& \textbf{SMART (Ours)} & \textbf{92.99} & \textbf{92.64} & \textbf{90.40} & \textbf{90.17} & \textbf{0.64} & \textbf{98.10} & \textbf{97.95} & \textbf{96.72} & \textbf{91.96} & \textbf{0.48} \\

\midrule

\multirow{8}{*}{DS KSL}& SMART (CSLR-Only) & 53.25 & 38.54 & 5.50 & 62.40 & 27.27 & 53.81 & 37.15 & 5.90 & 60.36 & 23.92 \\
& HS-I3D~\cite{eccv2022_hierarchical}& 23.40& 19.56& 3.11& 33.43& -& 29.88& 20.75& 3.34& 34.23& -\\
& ASFormer~\cite{sjh-asformer21} & 55.20 & 52.04 & 40.40 & 67.54 & 52.41 & 61.71 & 58.75 & 45.75 & 69.19 & 43.46 \\
& ASRF~\cite{sjh-asrf21} & 48.81 & 44.74 & 32.43 & 66.67 & 62.78 & 59.39 & 55.41 & 40.57 & 68.34 & 50.57 \\
& LTContext~\cite{sjh-ltcontext23} & 61.20 & 54.10 & 38.50 & 65.40 & 61.42 & 64.80 & 57.70 & 41.11 & 65.71 & 60.30 \\
& VAC + CSFormer & 69.87 & 66.64 & 52.69 & 72.42 & 30.83 & 74.56 & 71.25 & 57.17 & 73.68 & 26.68 \\
& SEN + CSFormer & \underline{72.28}& \underline{69.30}& 56.33& \underline{74.49}& 28.33& \underline{76.29}& \underline{73.52}& \underline{59.22}& \underline{74.72}& 24.55\\
& CorrNet + CSFormer & 72.23 & 69.08 & 55.16 & 74.03 & \underline{26.66}& 76.13 & 73.16 & 58.35 & 74.23 & 23.49 \\
& AdaptSign + CSFormer & 70.48 & 69.12 & \underline{56.40} & 74.14 & 26.75 & 74.67 & 72.73 & 58.16 & 74.57 & \underline{23.23}\\
\cmidrule(lr){2-12}

& \textbf{SMART (Ours)} & \textbf{72.45} & \textbf{69.53} & \textbf{56.51} & \textbf{74.51} & \textbf{26.34} & \textbf{76.31} & \textbf{73.56} & \textbf{59.77} & \textbf{74.77} & \textbf{22.93} \\

\bottomrule
\end{tabular}
}
\end{table*}
\subsection{Comparison with State-of-the-Art}

\noindent\textbf{Sign Language Recognition Results.}
Tab.~\ref{tab:main_recognition} compares \textbf{SMART} with state-of-the-art methods 
on four benchmarks: PHOENIX14-T, CSL-Daily, Large-scale KSL and DS KSL. \textbf{SMART} achieves the best performance across all datasets, outperforming previous methods in terms of WER. These results show the effectiveness of the proposed MLLM-guided video-text alignment and MSTA for CSLR. 

\noindent\textbf{Comparison with Segmentation and Spotting Baselines.} 
As shown in Tab.~\ref{tab:spotting_main_both}, we evaluate \textbf{SMART} against conventional action segmentation methods and a sign spotting baseline~\cite{eccv2022_hierarchical}. On Large-scale KSL, the spotting baseline improves over CSLR-only at low IoU thresholds, but does not improve boundary localization at a higher IoU threshold; for example, Test F1@50 changes from 7.37 to 7.13. On DS KSL, the same baseline performs even worse than CSLR-only across all IoU thresholds. This indicates that sparse query-based spotting methods are not well suited for dense frame-level spotting in continuous sign videos. Although action segmentation methods achieve stronger localization, they suffer from significantly higher WER, indicating that visual temporal boundaries alone are insufficient without gloss-level recognition guidance. In contrast, \textbf{SMART} achieves strong performance on both spotting and recognition, obtaining 96.72 F1@50 with 0.48 WER on Large-scale KSL and 59.77 F1@50 with 22.93 WER on DS KSL. 

To further validate CSFormer, we integrate it into existing CSLR models (e.g., VAC, SEN, CorrNet, and AdaptSign). With CSFormer, CSLR models consistently obtain dense spotting predictions, showing that sparse recognition cues can be transformed into frame-level localization. In particular, \textbf{SMART} improves Test F1@50 from 7.37 to 96.72 on Large-scale KSL while preserving WER and from 5.90 to 59.77 on DS KSL while reducing WER from 23.92 to 22.93. These results show that CSFormer bridges CSLR and sign spotting through gloss-level recognition guidance and dense temporal supervision.

\begin{table}[t]
\centering

\begin{minipage}[t]{0.42\linewidth}
\centering
\caption{Ablation study of \textbf{SMART} components.}
\label{tab:recognition_ablation}
\small
\setlength{\tabcolsep}{4pt}

\begin{tabular}{ccc cc}
\toprule
Fusion & Align & MSTA & Dev(\%) & Test(\%) \\
\midrule
-- & -- & -- & 18.60 & 19.80 \\
-- & -- & \checkmark & 17.88 & 19.82 \\
-- & \checkmark & -- & 17.98 & 19.75 \\
\checkmark & -- & \checkmark & 18.25 & 20.31 \\
\checkmark & \checkmark & \checkmark & 19.05 & 21.08 \\
-- & \checkmark & \checkmark & \textbf{17.58} & \textbf{19.50} \\
\bottomrule
\end{tabular}
\end{minipage}
\hspace{0.015\linewidth}
\begin{minipage}[t]{0.55\linewidth}
\centering
\caption{Ablation study of PEFT configurations.}  
\label{tab:peft}
\small
\setlength{\tabcolsep}{4pt}

\begin{tabular}{l c c c c}
\toprule
Method & Layers & Rank & Dev(\%) & Test(\%) \\
\midrule

\multirow{6}{*}{Adapter}
& 11--12 & 16 & 18.84 & 20.59 \\
& 11--12 & 32 & 18.28 & 20.47 \\

& 9--12 & 16 & 18.62 & 20.33 \\
& 9--12 & 32 & \textbf{17.58} & 19.50 \\

& 5--12 & 16 & 18.09 & 20.05 \\
& 5--12 & 32 & 17.88 & \textbf{19.39} \\

\midrule
LoRA & 9--12 & 32 & 24.40 & 25.20 \\

\bottomrule
\end{tabular}
\end{minipage}

\end{table}

\begin{table}[t]
\centering
\caption{Effect of batch size and video-text alignment methods.}
\label{tab:alignment}
\small
\begin{tabular}{c cc | c cc} 
\toprule
Batch Size & Dev(\%) & Test(\%) & Alignment & Dev(\%) & Test(\%) \\
\midrule
2 & 19.13 & 20.36 & None    & 18.60 & 19.80 \\
4 & 18.41 & 20.64 & InfoNCE & 18.62 & 20.76 \\
8 & \textbf{17.58} & \textbf{19.50} & SigLIP  & \textbf{17.98} & \textbf{19.75} \\
\bottomrule 
\end{tabular}
\end{table} 

\vspace{1mm}


\subsection{Ablation study}
\noindent\textbf{Analysis of each module in \textbf{SMART}.} 

\noindent\textit{Sign Language Recognition.}
Tab.~\ref{tab:recognition_ablation} reports the component ablation of the recognition module on the PHOENIX14-T dataset. Compared with the baseline, MSTA reduces the Dev WER from 18.60 to 17.88, showing the effectiveness of MSTA. The alignment objective also improves recognition performance, reducing the Test WER from 19.80 to 19.75. By jointly applying MSTA and alignment, the model achieves the best performance, with 17.58 Dev WER and 19.50 Test WER. In contrast, using MLLM-generated text features via a gated fusion module leads to performance degradation, indicating that the generated descriptions are more effective as auxiliary semantic supervision than as input features for recognition. 

\begin{table}[t]
\centering
\caption{Ablation study of CSFormer on Large-scale KSL and DS KSL. F1@50 and WER are reported to evaluate sign spotting and spotting-enhanced recognition, respectively.}
\label{tab:spotting_ablation}
\scriptsize
\resizebox{\textwidth}{!}{%
\begin{tabular}{ccc cc cc cc cc}
\toprule
& & & \multicolumn{4}{c}{Large-scale KSL} & \multicolumn{4}{c}{DS KSL} \\
\cmidrule(lr){4-7} \cmidrule(lr){8-11}

& & & \multicolumn{2}{c}{Dev} & \multicolumn{2}{c}{Test}
      & \multicolumn{2}{c}{Dev} & \multicolumn{2}{c}{Test} \\
\cmidrule(lr){4-5} \cmidrule(lr){6-7}
\cmidrule(lr){8-9} \cmidrule(lr){10-11}

Bnd head& Xattn& CSLR feat& F1@50 & WER(\%)& F1@50 & WER(\%)& F1@50 & WER(\%)& F1@50 & WER(\%)\\
\midrule
            --& --& --& 87.42& -& 94.50& -& 54.25& 27.13& 58.54& 23.72\\
\checkmark&       --& --& 88.24& -& 94.92& -& 54.09& 27.07& 57.96& 23.74\\
     --&  --& \checkmark& 88.80& -& 95.44& -& 56.50& 26.92& 59.68& 23.68\\
\checkmark& --& \checkmark& 89.58& -& 94.75& -& 56.45& 26.95& 58.78& 23.65\\
--& \checkmark& \checkmark& 90.35& -& \textbf{97.06}& -& 56.43& 27.05& \textbf{59.81}& 23.66\\
 \checkmark& \checkmark& \checkmark& \textbf{90.40}& 0.64& 96.72& 0.48& \textbf{56.51}& \textbf{26.34}& 59.77&\textbf{22.93}\\
\bottomrule
\end{tabular}%
}
\end{table}

\noindent\textit{Sign Language Spotting.}
Tab.~\ref{tab:spotting_ablation} reports the ablation study on the LS KSL and DS KSL datasets to validate the proposed CSFormer. The baseline without CSLR-derived guidance shows limited spotting performance, while adding CSLR features improves Test F1@50 from 94.50 to 95.44 on Large-scale KSL and from 58.54 to 59.68 on DS KSL. This confirms that recognition-derived gloss probabilities provide effective gloss-level guidance for sign spotting. Furthermore, combining CSLR features with cross-attention achieves the strongest spotting performance, reaching 97.06 Test F1@50 on Large-scale KSL and 59.81 on DS KSL, indicating that layer-wise interaction between the visual and CSLR streams is beneficial for dense temporal localization. Although the full model slightly sacrifices F1@50 compared with the best partial variant, it maintains competitive spotting performance and achieves the lowest WER on DS KSL, reducing Test WER from 23.72 to 22.93. Therefore, we use the full CSFormer as the default configuration for balancing sign spotting and sequence-level recognition.

\noindent\textbf{Parameter-Efficient Fine-Tuning (PEFT).}
Tab.~\ref{tab:peft} evaluates MSTA applied to the last $N \in \{2, 4, 8\}$ transformer layers with reduction factors $r \in \{16, 32\}$ on PHOENIX14-T. Additional results with smaller reduction factors are provided in the supplementary material. Applying MSTA to the last 4 layers with $r=32$ achieves the best Dev WER of 17.58. Although applying MSTA to the last 8 layers improves Test WER from 19.50 to 19.39, it doubles the adapter parameters and degrades Dev WER from 17.58 to 17.88. Therefore, we adopt the last 4 layers with $r=32$ for parameter-efficient adaptation. Under the same configuration, LoRA performs worse, with 24.40 Dev and 25.20 Test WER.


\noindent\textbf{Video-Text Alignment Analysis.}
Tab.~\ref{tab:alignment} analyzes the effect of batch size and alignment method on PHOENIX14-T. Increasing the batch size generally improves recognition performance, with batch size 8 achieving the best results of 17.58 Dev WER and 19.50 Test WER. Since larger batch sizes exceed GPU memory capacity for sign language video training, batch size 8 is the practical upper bound in our setting. For the alignment objective, we use the same baseline model~\cite{hu2024improving} without MSTA and vary only the alignment objective. InfoNCE~\cite{radford2021learning} performs worse than the no-alignment baseline, increasing Test WER from 19.80 to 20.76, while SigLIP~\cite{zhai2023sigmoid} improves it to 19.75.


\subsection{Qualitative Analysis}


\begin{figure*}[t]
    \centering

    \begin{minipage}{0.98\textwidth}
        \centering
        \includegraphics[width=\textwidth]{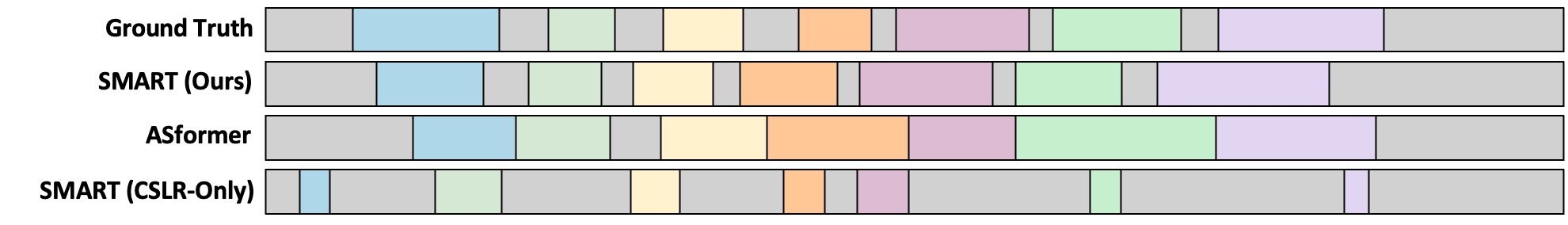}
        \vspace{1mm}
        \small (a) Large-scale KSL dataset.
    \end{minipage}

    \vspace{1mm}

    \begin{minipage}{0.98\textwidth}
        \centering
        \includegraphics[width=\textwidth]{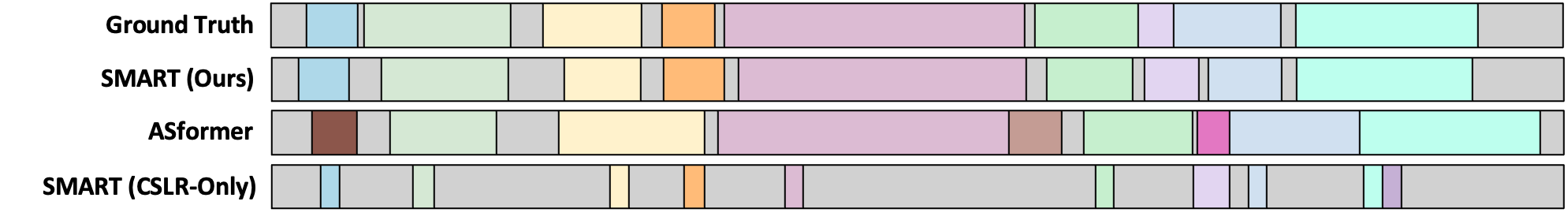}
        \vspace{1mm}
        \small (b) DS KSL dataset.
    \end{minipage}

    \caption{
    Qualitative comparison of sign spotting performance on (a) Large-scale KSL and (b) DS KSL datasets. Different colors denote distinct glosses, while the block widths correspond to their temporal durations.
    }
    \label{fig:spotting_result}
\end{figure*}


\begin{figure*}[t] 
    \centering
    \includegraphics[width=0.9\textwidth]{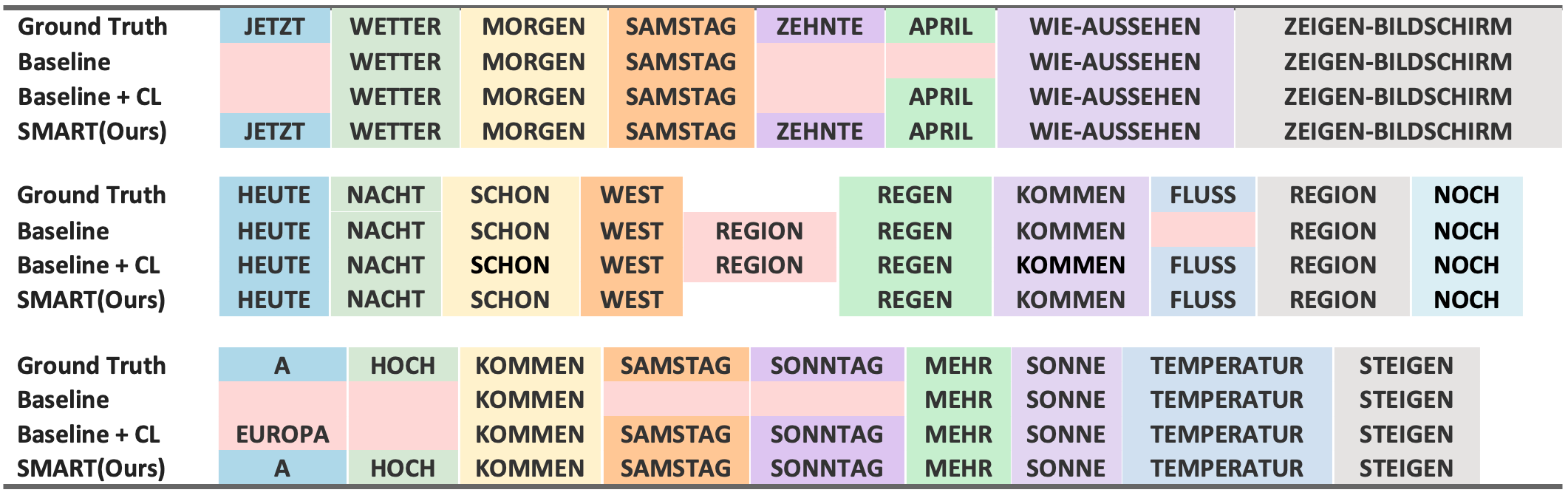}
    \vspace{2mm}
    \caption{Gloss prediction comparison between the baseline model, the baseline model with contrastive learning (CL), the proposed \textbf{SMART} framework, and ground truth annotations. Incorrect gloss predictions are highlighted in pink.
    } 
    \label{fig:gloss}
\end{figure*}

\begin{figure*}[t]
\centering
\scriptsize
\setlength{\tabcolsep}{1.5pt}
\renewcommand{\arraystretch}{1.1}

\begin{tabular}{lcccccc}
\toprule

&
\multicolumn{3}{c}{\textbf{Dev}}
&
\multicolumn{3}{c}{\textbf{Test}} \\

\cmidrule(lr){2-4}
\cmidrule(lr){5-7}

\textbf{Dataset}
& \textbf{Original}
& \textbf{Baseline}
& \textbf{SMART}
& \textbf{Original}
& \textbf{Baseline}
& \textbf{SMART} \\

\midrule

\raisebox{2.8em}{\textbf{PHOENIX14-T}}
&
\includegraphics[width=0.12\textwidth]{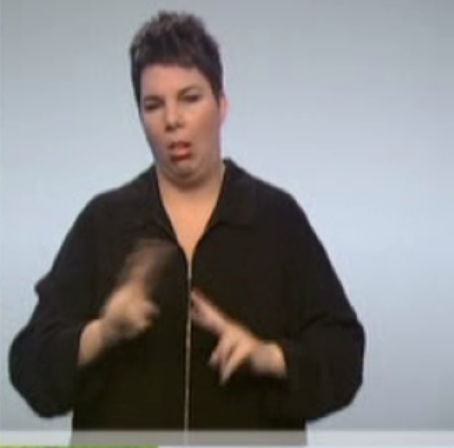}
&
\includegraphics[width=0.12\textwidth]{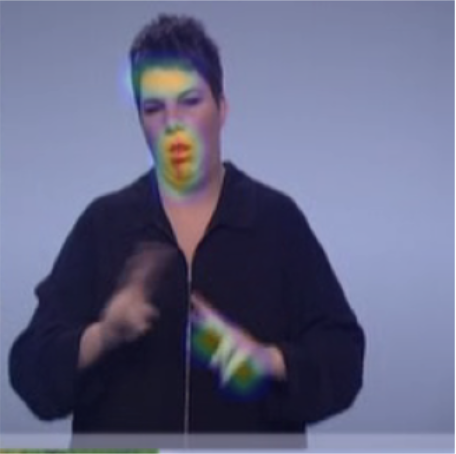}
&
\includegraphics[width=0.12\textwidth]{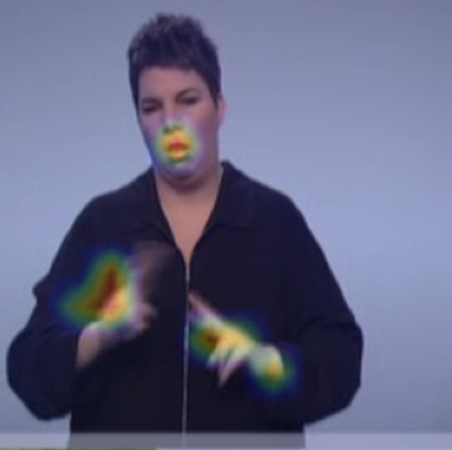}
&
\includegraphics[width=0.12\textwidth]{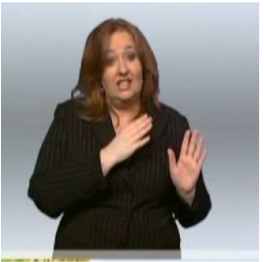}
&
\includegraphics[width=0.12\textwidth]{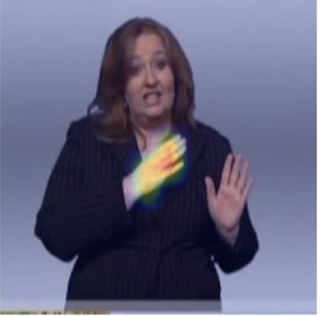}
&
\includegraphics[width=0.12\textwidth]{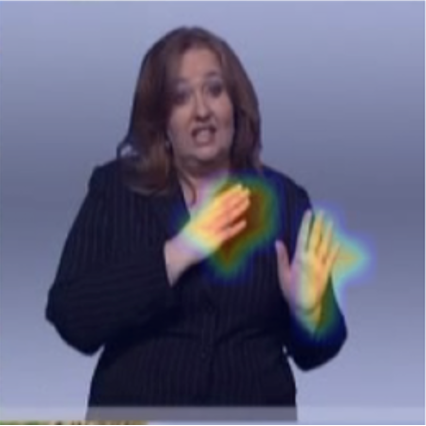}
\\

\raisebox{2.8em}{\textbf{CSL-Daily}}
&
\includegraphics[width=0.12\textwidth]{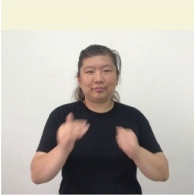}
&
\includegraphics[width=0.12\textwidth]{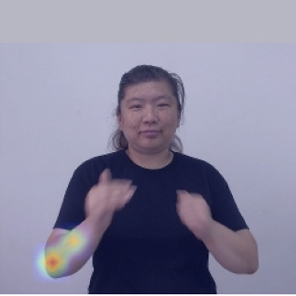}
&
\includegraphics[width=0.12\textwidth]{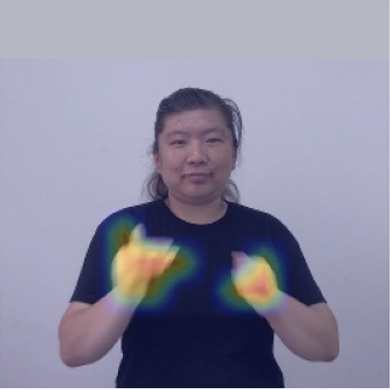}
&
\includegraphics[width=0.12\textwidth]{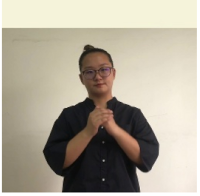}
&
\includegraphics[width=0.12\textwidth]{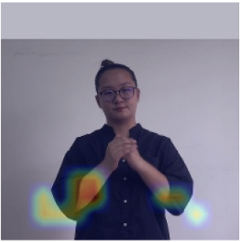}
&
\includegraphics[width=0.12\textwidth]{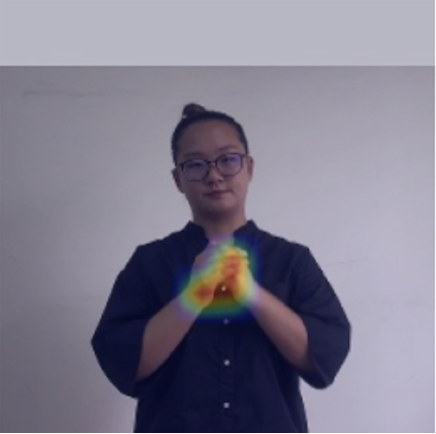}
\\

\raisebox{2.8em}{\textbf{Large-scale KSL}}
&
\includegraphics[width=0.12\textwidth]{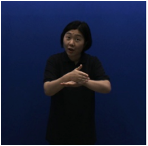}
&
\includegraphics[width=0.12\textwidth]{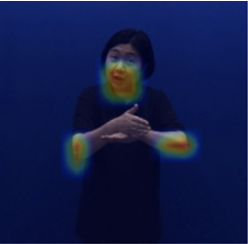}
&
\includegraphics[width=0.12\textwidth]{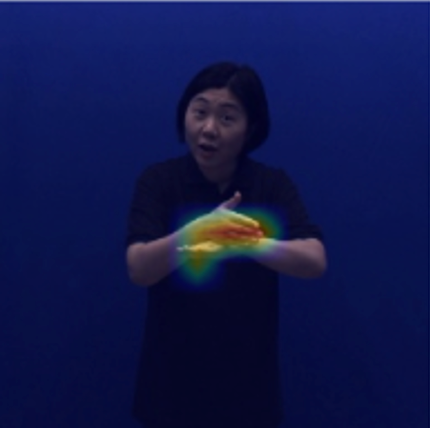}
&
\includegraphics[width=0.12\textwidth]{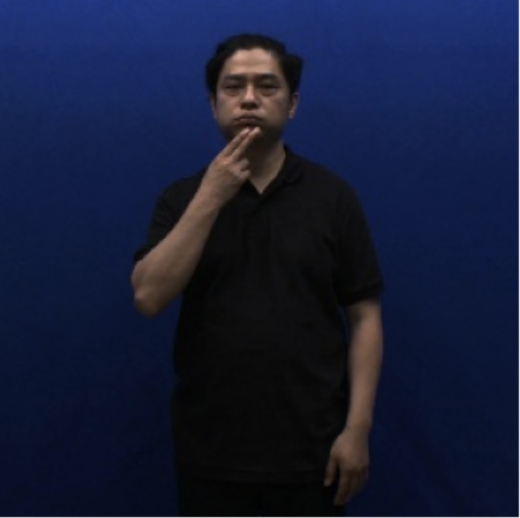}
&
\includegraphics[width=0.12\textwidth]{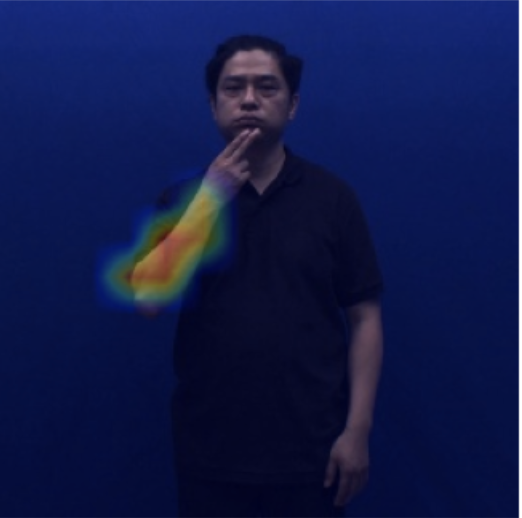}
&
\includegraphics[width=0.12\textwidth]{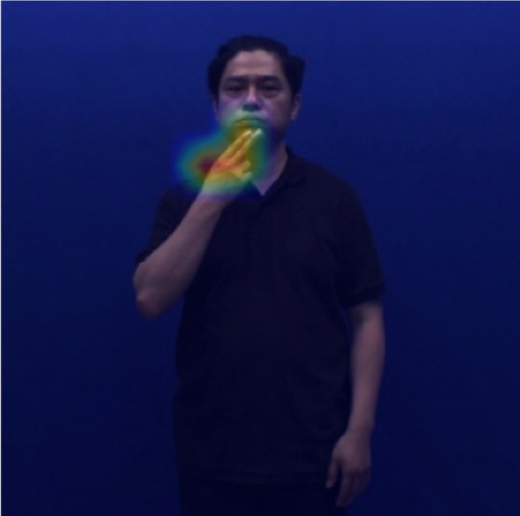}
\\

\raisebox{2.8em}{\textbf{DS KSL}}
&
\includegraphics[width=0.12\textwidth]{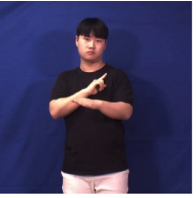}
&
\includegraphics[width=0.12\textwidth]{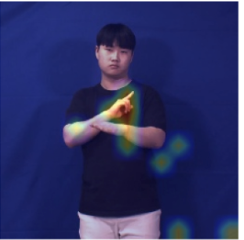}
&
\includegraphics[width=0.12\textwidth]{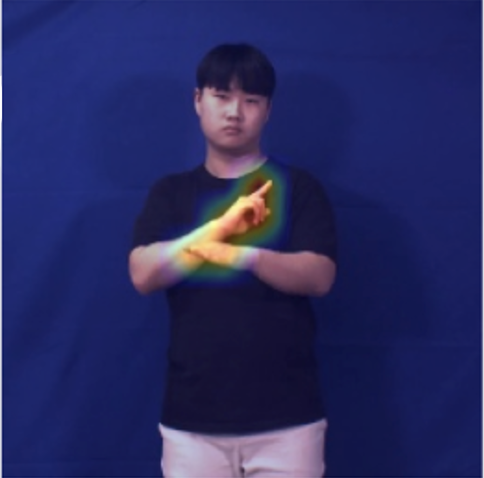}
&
\includegraphics[width=0.12\textwidth]{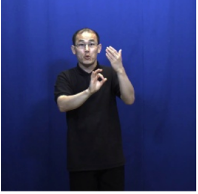}
&
\includegraphics[width=0.12\textwidth]{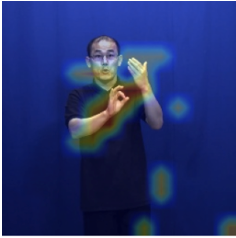}
&
\includegraphics[width=0.12\textwidth]{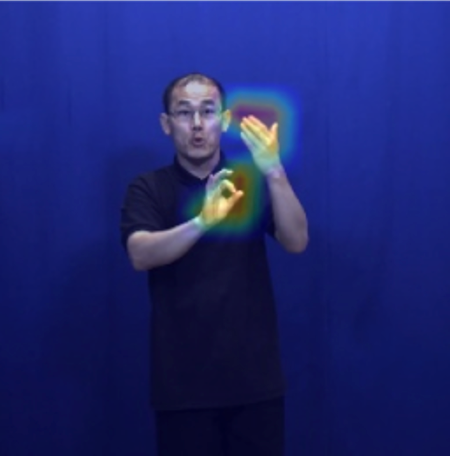}
\\

\bottomrule
\end{tabular}

\caption{Qualitative comparison of attention maps between the baseline and \textbf{SMART} on dev and test samples across four CSLR benchmarks.
}

\label{fig:attention_map}
\end{figure*}

\noindent\textbf{Qualitative results on Sign Spotting.} Fig.~\ref{fig:spotting_result} visualizes spotting results on the Large-scale KSL and DS KSL datasets. \textbf{SMART} (CSLR-Only) exhibits typical peaky alignments: it captures gloss identities at sparse time steps, but fails to cover the full temporal duration of each sign. Conversely, the baseline segmentation method, ASFormer~\cite{sjh-asformer21}, generates dense segments but lacks gloss-level discrimination, suffering from frequent misclassifications and merging adjacent signs. By injecting CSLR-derived gloss guidance into CSFormer, our unified \textbf{SMART} framework overcomes these limitations and produces predictions that better match the ground truth in both temporal boundaries and gloss labels. 

\noindent\textbf{Visualization of Gloss Prediction Results.}
Fig.~\ref{fig:gloss} shows qualitative gloss prediction results on PHOENIX14-T. The baseline model~\cite{hu2024improving} shows frequent deletion and substitution errors, particularly for less frequent signs. Adding contrastive learning (CL) to the baseline improves predictions, but deletion and substitution errors remain in several cases. For example, \textbf{SMART} correctly predicts JETZT and ZEHNTE in the first sequence and A and HOCH in the third, which are missed by the baseline. Overall, \textbf{SMART} achieves predictions closer to the ground truth with fewer deletion and substitution errors.


\noindent\textbf{Qualitative Analysis of Attention Maps.} 
Fig.~\ref{fig:attention_map} shows attention visualizations on four CSLR benchmarks using samples from the dev and test sets. Compared with the baseline~\cite{hu2024improving}, \textbf{SMART} focuses more on hand movements and facial expressions with less attention to the background. These results suggest that MLLM-guided SigLIP alignment and MSTA improve spatial and temporal feature representations.

\section{Conclusions}
\label{sec:conclusion}
In this work, we proposed \textbf{SMART}, an MLLM-guided Temporal Alignment framework for sign language recognition and spotting. \textbf{SMART} introduces video-text alignment with MLLM-generated motion descriptions, temporal representation learning with a Multi-Scale Temporal Adapter (MSTA), and CSLR-guided sign spotting via CSFormer for temporal boundary refinement. CSFormer injects recognition-derived gloss probabilities into a boundary-aware spotting module to produce dense frame-level predictions for accurate sign localization. Experiments on four sign language benchmarks show that \textbf{SMART} achieves state-of-the-art performance across sign languages. Additional spotting experiments show that CSLR representations provide effective gloss-level guidance for sign spotting, while dense spotting supervision improves or preserves recognition performance depending on dataset characteristics. These results indicate that gloss-level recognition guidance and dense temporal supervision are complementary for improving continuous sign language understanding.


\section{Acknowledgement}
\label{sec:acknowledgement}
This work was supported by the Institute of Information \& Communications Technology Planning \& Evaluation (IITP), funded by the Korean government (Ministry of Science and ICT), through the ITRC (IITP-2026-RS-2024-00437102, 50\%) and ICAN (IITP-2026-RS-2024-00437027, 50\%) programs.



\bibliography{egbib}
\end{document}